\documentclass[runningheads]{llncs}
\usepackage{array,multirow,graphicx}
\usepackage{amssymb}
\begin{document}
\title{A compact sequence encoding scheme for online human activity recognition in HRI applications}
%
%
\author{Georgios Tsatiris\inst{1} \and
Kostas Karpouzis\inst{1} \and
Stefanos Kollias\inst{1,2}}
\authorrunning{G. Tsatiris et al.}
%
\institute{Artificial Intelligence and Learning Systems Laboratory, School of Electrical and Computer Engineering, National Technical University of Athens, 9 Heroon Politechniou str., 15780, Athens, Greece\and
School of Computer Science, University of Lincoln, Brayford Pool, Lincoln, Lincolnshire, UK\\
\email{gtsatiris@image.ntua.gr, kkarpou@cs.ntua.gr, stefanos@cs.ntua.gr}}
\maketitle              
\begin{abstract}
Human activity recognition and analysis has always been one of the most active areas of pattern recognition and machine intelligence, with applications in various fields, including but not limited to exertion games, surveillance, sports analytics and healthcare. Especially in Human-Robot Interaction, human activity understanding plays a crucial role as household robotic assistants are a trend of the near future. However, state-of-the-art infrastructures that can support complex machine intelligence tasks are not always available, and may not be for the average consumer, as robotic hardware is expensive. In this paper we propose a novel action sequence encoding scheme which efficiently transforms spatio-temporal action sequences into compact representations, using Mahalanobis distance-based shape features and the Radon transform. This representation can be used as input for a lightweight convolutional neural network. Experiments show that the proposed pipeline, when based on state-of-the-art human pose estimation techniques, can provide a robust end-to-end online action recognition scheme, deployable on hardware lacking extreme computing capabilities.

\end{abstract}
\section{Introduction}
Human activity recognition is a very important field of study inside the broad domain of pattern recognition, with many everyday life applications. High level human actions are an important information modality for social interaction and they play a critical role in Human-Robot Interaction (HRI) \cite{Ji2019}. Human poses, hand gestures and overall body motion are also important for human-focused HRI, especially in applications with robots interacting with humans daily (e.g. \cite{malatesta2009emotion}, \cite{bevacqua2006multimodal} and \cite{peters2011fundamentals}). Robotic assistants are required to understand and analyse human activity at all levels of interpretation, as well as have the ability to predict intentions and to imitate human actions when forming responses. Social robots and applications aimed at such environments will need to be employed in human-centric contexts, build trust and effectively assist humans beings \cite{Chen2018}. In this paper, we focus on online human activity recognition pipelines with the aim to provide solutions on the inherent problems of such endeavours.

\subsection{Related Work}
Various works on activity analysis and understanding deal with real application scenarios. In the field of preventing unintentional falls in elderly care and assistive environments, the work presented in \cite{Yu2017} is one of the most typical attempts for applying convolutional neural networks in a fall detection application. A dataset of daily activities recorded from many people is used as classification is performed on single frames containing human postures. This method performs well on posture classification and can detect falls with a low false alarm rate. In an earlier work \cite{Liu2016}, an efficient algorithm is proposed to identify temporal patterns in actions and construct activity representations for automated recognition.

There is an extensive literature of studies focused on HRI, a comprehensive presentation of which can be found in \cite{Ji2019}. In a typical example of such works, authors in \cite{Rodomagoulakis2016} propose that state-of-the-art approaches from automatic speech and visual action recognition, fused with multiple modalities, can construct multimodal recognition pipelines focused on assistive robots for elderly environments. An end-to-end approach shown in \cite{Gerard2016} deals with dynamic gestures using a Dynamic Time Warping approach, based on features extracted from depth data, in a complete HRI scenario. 

\subsection{Real world challenges}

As stated in \cite{Tsatiris2018}, applied human activity recognition pipelines are demanding in terms of hardware and general infrastructure. Most action recognition systems, especially smart-home or assistive living applications, depend on network infrastructures for easy data fusion and integration of different sensing modalities. HRI environments are no different, in the sense that they may need computing capabilities similar to common workstations, as well as networking. Especially when dealing with deep learning-based pipelines, we are used to having hardware acceleration (i.e. massive parallel computing, GPU acceleration etc.) at our disposal. This study aims to provide a lightweight pipeline with limited need for high-end infrastructures.

\section{Theoretical Primer}

In the following section, we attempt a brief documentation of the theoretical background on which the proposed scheme is built upon. This primer emphasizes on the statistical nature of spatio-temporal data, the use of the Radon Transform for feature encoding, as well as the necessary tools to encode the necessary information into a compact pipeline.

\subsection{Variance-based Features}
As mentioned in \cite{tsatiris2017}, high level human pose and activity descriptors, such as spatio-temporal interest points \cite{laptev2005} or human joints, can be treated as probabilistic distributions. Joints detected in a video frame or still image, e.g. using pipelines such as OpenPose \cite{cao2018}, can be considered distributed in space, whereas spatio-temporal interest points are distributed both in space and time. Action recognition techniques benefit from accurately modeling these relationships between the salient features describing an action sequence.

Particularly in \cite{tsatiris2017}, a method to distinguish between experienced tennis players and amateurs is shown, which employees feature vectors constructed by calculating the per frame variance of interest points. Variance, as the second central moment, is an effective shape descriptor \cite{Loncaric1998}. Interest points and joint locations form a 3D volumes in space and time. In that context, we extend the idea demonstrated in \cite{tsatiris2017}, by incorporating the use of Mahalanobis distances.

\subsubsection{Mahalanobis Distance}

Given a sample vector $\vec{x}$ from a distribution $D$ with mean $\vec{\mu}$ and covariance matrix $S$, the Mahalanobis distance between $\vec{x}$ and $D$ is given by equation \ref{mahal}.

\begin{equation}
mahal(\vec{x}, \vec{\mu}, S) = \sqrt{(\vec{x} - \vec{\mu})^TS^{-1}(\vec{x} - \vec{\mu})}
\label{mahal}
\end{equation}

In this work, we are focusing on detected human joints in 3D space, like the ones extracted using depth sensors (e.g. Intel RealSense\footnote{https://www.intelrealsense.com/} or Microsoft Azure Kinect\footnote{https://azure.microsoft.com/en-us/services/kinect-dk/}) and by pipelines such as the one presented in \cite{Nibali2019}, which extends common 2D heatmap-based pose detection techniques such as \cite{cao2018} without utilizing costly volumetric operations. There is a good indication that skeletal joints can be considered valid spatio-temporal interest points when it comes to human action analysis and, with them being a more refined and targeted source of human posture information, can lead to better results \cite{varia2018}.

As the relative position of a joint and the human body is crucial in the definition of human motion, we believe that a compact and robust way to model these individual relations should be central to any action recognition pipeline. The Mahalanobis distance, as an established metric for point-to-distribution distances, is more suitable in this task, instead of more intuitive metrics, such the Euclidean distance between a point and the distribution centroid. Its place in this pipeline will be explained in detail in subsection \ref{sec_srf}.

\subsection{Use of the Radon Transform}
The Radon transform is effectively used as a reconstruction and feature extraction tool for several decades \cite{Beylkin1987}. It has found applications in earlier human activity recognition pipelines based on hand-crafted features \cite{chen2008} and in novel techniques, alongside its generalization, the Trace transform \cite{kadyrov2001}, in offline pipelines \cite{Goudelis2013}\cite{goudelis2017}. Another application of the Radon transform in this field leverages the transform's ability to create translation and scale invariant features to use as input for a Recurrent Neural Network (RNN) \cite{Uddin2018}. However, for reasons which will be demonstrated in the next section, this paper avoids the use of RNN and draws inspiration from such studies as the one in \cite{Hamdi2018}, where source images and their Conic Radon transforms are fed into a CNN for fingerprint classification.

Given a 2D function $f(x,y)$ with $x, y \in \mathbb{R}$, its Radon transform $R_f$ is equal to the integral of all values of $f$ under all lines $L$, defined by parameters $\rho,\theta \in \mathbb{R}$, where $\rho$ is each line's distance from the origin and $\theta$ is the angle formed between the line and the horizontal axis:

\begin{equation}
R_f(\rho,\theta) = \int_{L} f(x,y) dL
\label{radon1}
\end{equation}

\noindent If we substitute the parametrical $\rho,\theta$ line equation in equation \ref{radon1}, we have the following: 

\begin{equation}
R_f(\rho,\theta) = \int_{-\infty}^{\infty}\int_{-\infty}^{\infty} f(x,y)\delta(x\cos\theta + y\sin\theta - \rho) dxdy
\label{radon2}
\end{equation}

\noindent where $\delta$ the Dirac delta function, ensuring that only the values of $f$ under line $\rho,\theta$ will be integrated over.

\section{The Proposed Pipeline}

At this point, we have documented the theoretical tools to give spatio-temporal prominence to raw joint information using variance-based shape features and, particularly, the Mahalanobis distance. We have also seen how the Radon transform and other relevant transforms can create robust features out of otherwise non-salient 2D and 3D functions. This section will clarify how the theoretical tools will be used in a unified and compact pipeline action encoding pipeline.

\subsection{Spatio-temporal Radon Footprints}
\label{sec_srf}

As we have established, the relationship between the position of one joint and the rest of the body parts of a subject, at a particular point in time, can be encoded by treating the set of joints as a distribution and calculating the Mahalanobis distance (given by equation \ref{mahal}) between it and that particular joint. In our context, a point in time is a frame in an action sequence or video. So, if we extend the above claim to all the joints of a subject and across all frames of an action sequence, we can formulate a Mahalanobis matrix $M$, whose values for joint $j$ and frame $f$ can be calculated using the following equation:

\begin{equation}
M(j,f) = mahal(\vec{x^j_f}, \vec{\mu_f}, S_f)
\label{mahalMat}
\end{equation}

\noindent where $\vec{x^j_f}$ is the coordinate vector of joint $j$ in frame $f$, defined in $\mathbb{R}^2$ or $\mathbb{R}^3$, $\vec{\mu_f}$ is the mean of all joint coordinate vectors in frame $f$ (essentially the centroid) and $S_f$ is the covariance matrix of the joint distribution at frame $f$.

In essence, the Mahalanobis matrix formulated in equation \ref{mahalMat} is a complete 2D representation of the action sequence in space and time. With $J\times t$ resolution, where $J$ is the total number of joints and  $t$ is the point in time (number of frames passed) since the beginning of the sequence, each line of the matrix has the Mahalanobis distances of every joint at the corresponding frame.

What this representation lacks, however, is a way to correlate the individual frame information into a compact robust feature which will describe the action sequence from its start up to point $t$. For this reason, we calculate the Radon transform of the 2D function $M$ by substituting the representation $M(j,f)$ into equation \ref{radon2}. This gives as a 2D spatio-temporal feature describing the action sequence up to point $t$ in time. This feature will, for the rest of this paper, be called the Spatio-temporal Radon Footprint of the action up to point $t$, will be deonted as $SRF_t$ and is calculated as follows: 

\begin{equation}
SRF_t(\rho,\theta) = \int_{0}^{t-1}\int_{0}^{J - 1} M(j,f)\delta(j\cos\theta + f\sin\theta - \rho) djdf
\label{SRF}
\end{equation}

\noindent In the above equation, it must hold that $t \geqslant 2$, both for $M$ to be a 2D function of $j$ and $f$ and for the representation to encode a minimum of temporal information. In practice, depending on the action itself, more frames may need to pass before starting to calculate $SRF$s. Naturally, it should also hold that $F > 2$, which is obvious, both for the aforementioned reason as well as because it is impossible to ascertain the nature of an activity with insufficient joint information.

\begin{figure}
\includegraphics[width=\textwidth]{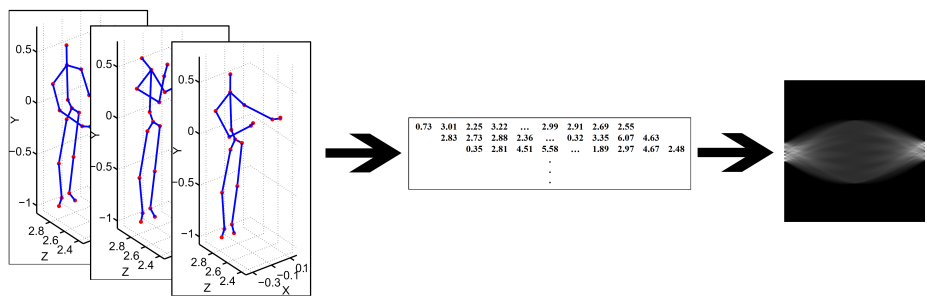}
\caption{Calculating $SRF_t$: The Mahalanobis matrix is calculated from the action sequence at point $t$ and the Radon transform of the matrix gives the final result.}
\label{srf_process}
\end{figure}

\begin{figure}
\includegraphics[width=\textwidth]{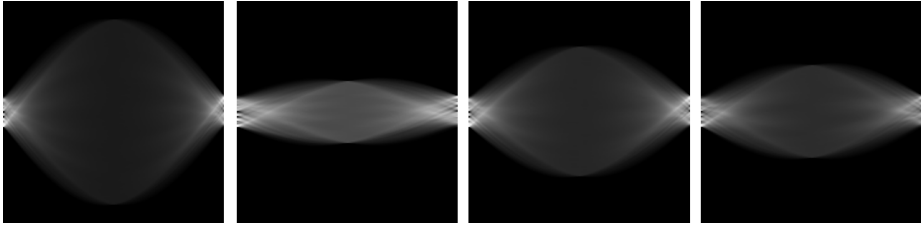}
\caption{$SRF$ samples from various action sequences of the UTD-MHAD dataset, at varying points in time.}
\label{srf_samples}
\end{figure}

The process of calculating a $SRF$ at a point in time, from a sequence of frames containing joint information, is depicted in figure \ref{srf_process}. The frames shown are samples from the UTD-MHAD dataset \cite{chen2015}, which will be documented in subsection \ref{dataset}. Figure \ref{srf_samples} shows sample $SRF$s taken from this dataset, from different sequences and at different timestamps.

\subsection{A VGG CNN-based pipeline}

\begin{figure}
\includegraphics[width=\textwidth]{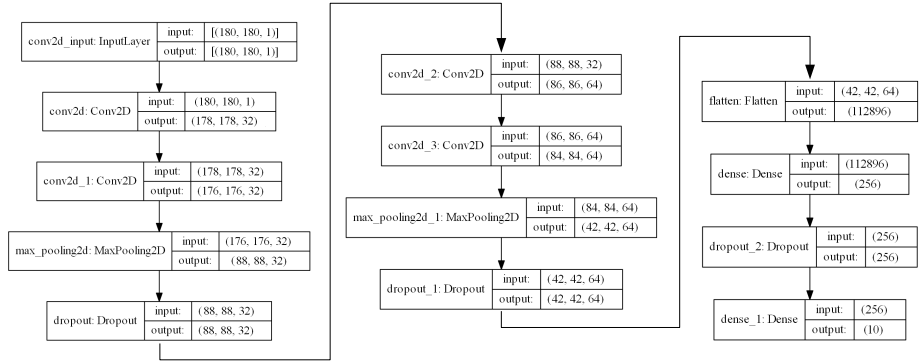}
\caption{The VGG-based architecture used in our experiments.} \label{vggnet}
\end{figure}

In previous works such as the ones in \cite{Goudelis2013} and \cite{goudelis2017}, the Radon transform and its derivatives were intermediate features, used to produce the final feature vector describing the action sequence, which in turn was used as input for established machine learning techniques, such as SVMs. In this work, however, much like in \cite{Hamdi2018}, instead of hand-crafting features based on the transform's result, a convolutional neural network will learn those features directly from the $SRF$s. Particularly, we opted for a VGG-based architecture \cite{Simonyan2015}, because of the simplicity, the ease to train and deploy and the ability to adapt to complex pattern recognition problems that this family of convolutional neural networks demonstrates. The network used in our pipeline is shown in figure \ref{vggnet}.

\subsubsection{Preference over RNN-LSTM architectures}
It is customary to use Recurrent Neural Network (RNN, LSTM) architectures for online pattern recognition tasks involving temporal sequences \cite{Liu2018}, due to their ability to learn and encode temporal information. However, known issues come with the use of such networks, including the need for large training datasets in most cases, the vanishing and exploding gradient problems, as well their high demands in computational resources \cite{Gehring2017}. By encoding online spatio-temporal information in $SRF$s and using robust object detection architectures, we aim to eliminate the need for such networks and provide an online and compact action recognition pipeline which would not depend on extremely sophisticated hardware.

\subsubsection{Per-frame and cumulative classification}

Human activity recognition is a temporally sensitive task. Offline techniques need to observe the complete action sequence before performing classification. In an online pipeline aiming towards real-time action recognition, per-frame classification is central. Previous studies, such as the one in \cite{Yu2017}, perform classification between unintentional falls and other activities on single postures/frames, rather than complete sequences. The methodology presented in this paper performs online action analysis, while taking into account the history of every previous per-frame classification task.

In a specific frame $f$ of the sequence, the $SRF_f$ is calculated and fed in the aforementioned network. The network then determines the class of the action at that frame. However, the final decision is made by counting the number each action class was deemed the correct class of the sequence, divided by the number of frames which have passed up until the moment of classification. In a sense, for each action class and in every frame, we calculate a confidence that this action belongs to that class. This confidence is updated per-frame, with the ultimate goal of determining the correct class as early in the sequence as possible. This way, we eliminate the possibility of outliers (frames being classified in the wrong class) among a number of correct classifications.

\section{Experimental Evaluation}

In this section, we discuss the experimental setup and the data used in the evaluation of the presented methodology. The choice of the validation protocol will also become apparent, as well as the efficiency of the proposed technique.

\subsection{Dataset}
\label{dataset}

\begin{figure}
\includegraphics[width=\textwidth]{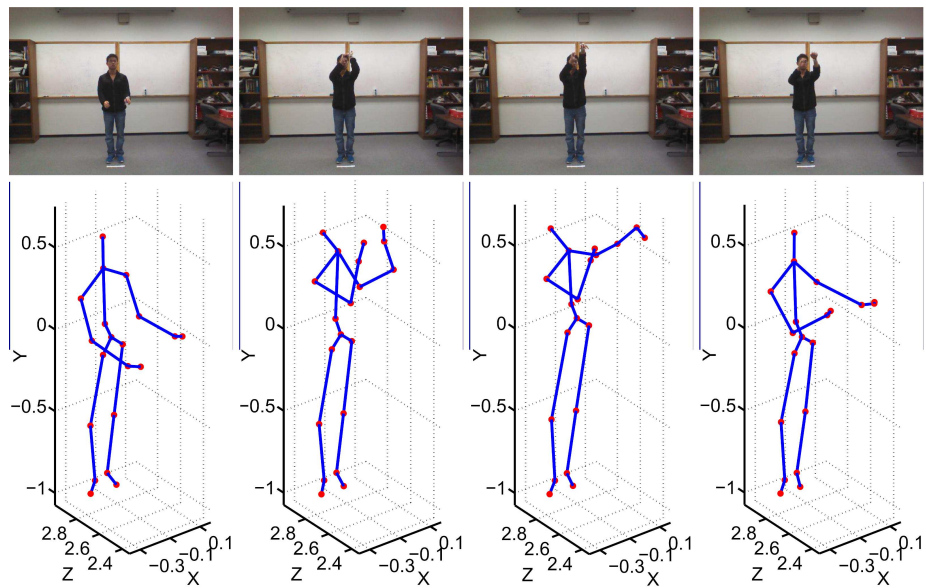}
\caption{Samples from the UTD-MHAD dataset.}
\label{data}
\end{figure}

In our experiments, we used the UTD-MHAD dataset \cite{chen2015}. In particular, we used a more recent subset of this dataset, which contains data captured using the Kinect v2 depth sensor. It includes 10 actions, namely "Right hand high wave", "Right hand catch", "Right hand high throw", "Right hand draw X", "Right hand draw tick", "Right hand draw circle", "Right hand horizontal wave", "Right hand forward punch", "Right hand hammer" and "Hand clap (two hands)". These actions were performed by 6 subjects (3 female and 3 male), each one of whom performed each action 5 times, resulting in a total of 300 action sequences. Figure \ref{data} shows sample data from the dataset.

We opted for this new subset of the UTD-MHAD dataset for two reasons, both of which are sensor related. The Kinect V2 is a relatively state of the art device which can detect joint information for 25 body joints without inducing much noise (Base of the spine, Middle of the spine, Neck, Head, Left shoulder, Left elbow, Left wrist, Left hand, Right shoulder, Right elbow, Right wrist, Right hand, Left hip, Left knee, Left ankle, Left foot, Right hip, Right knee, Right ankle, Right foot, Spine at the shoulder, Tip of the left hand, Left thumb, Tip of the right hand, Right thumb).

\subsection{Leave-One-Person-Out Cross-validation}

Validating a human activity recognition pipeline entails an inherent complexity which is not found in other classification tasks. In a typical scenario, we would have partitioned the dataset into a training and a testing set. Our dataset, as many other action datasets, is already partitioned by the subjects executing the actions. Simply leaving a person out for validation and training with the rest would not, however, produce reliable results.

Every person performs an action in a unique to every other person's way. For that matter, we perform the Leave-one-person-out cross-validation protocol (LOPO or LOO) across all individuals in our dataset. Essentially, we are training with the actions performed by 5 subjects and test with the actions performed by the one we left out. We repeat this process for all subjects in the dataset. The final reported accuracy would be the mean of the individual accuracies of each training-testing scenario.

\subsection{Results}

One of the first findings that we should report is that our pipeline achieved high training accuracy across all individual subject training tasks, with the mean training accuracy reaching \textbf{99.15\%}. The highest achieved training accuracy was as high as \textbf{99.25\%}, when training with all subjects except for subject \#3, which was used for testing.

An expected discrepancy was noticed across the validation tasks. Notably, when testing with subject \#6, the pipeline achieved validation accuracy equal to \textbf{90.08\%}, whereas the worst performance was reported when testing with with subject \#3 (\textbf{66.02\%}), on the same task that achieved the highest training accuracy. This hints towards possible issues of overfitting which may need to be dealt with in future works. The mean validation accuracy achieved after performing the complete LOPO protocol across all subjects was equal to \textbf{81.06\%}. Table \ref{tab1} shows the confusion matrix of the proposed pipeline after the validation process.

\begin{figure}
\includegraphics[width=\textwidth]{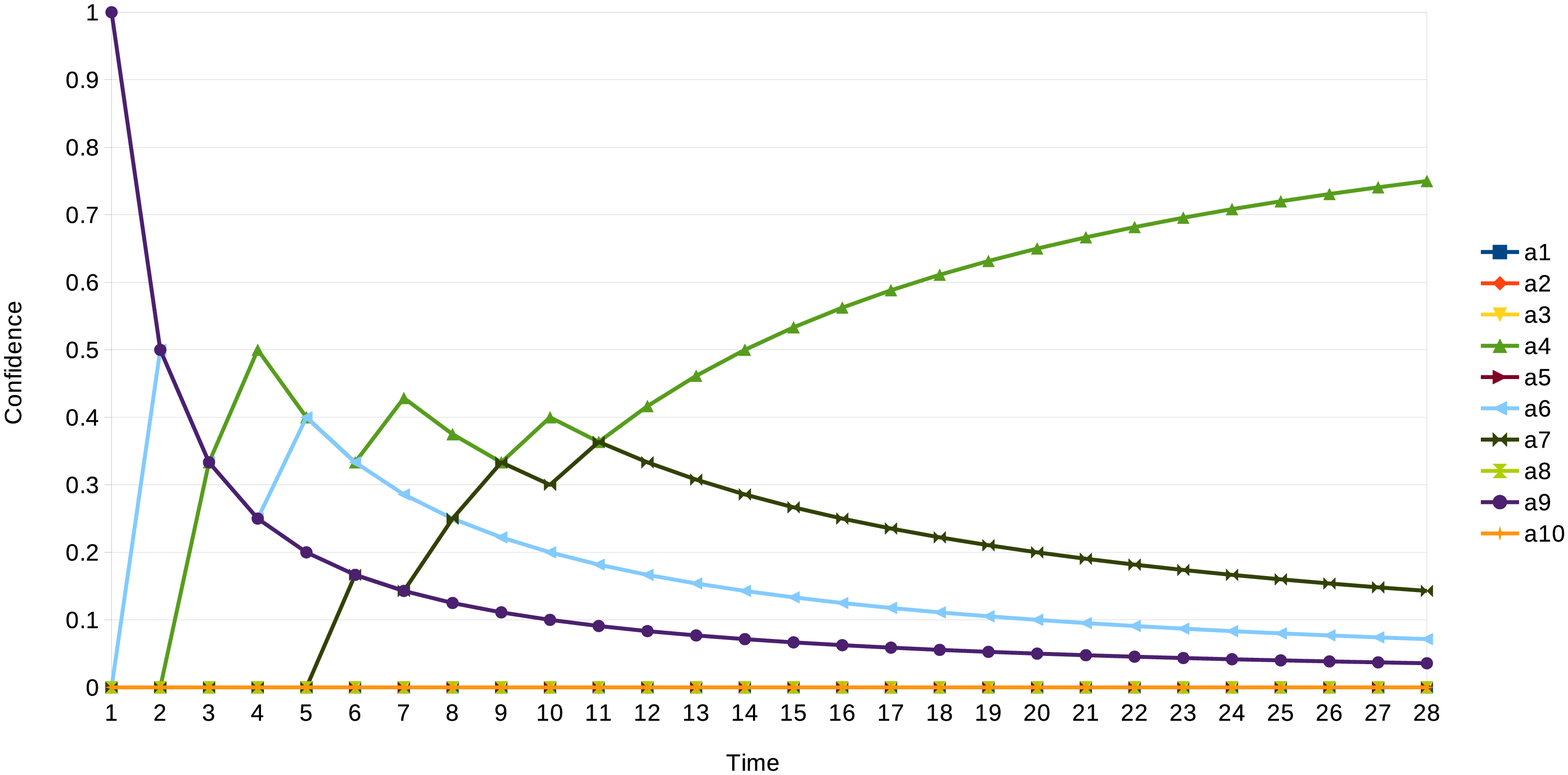}
\caption{Evolution of class confidences for action \#4, performed by subject \#6. After a certain time, correct class (4, Right hand draw X) emerges as most prevalent.}
\label{eval}
\end{figure}

Figure \ref{eval} shows a characteristic sample of the evolution of class confidences according to our proposed pipeline. In this figure we observe how action class 4 ("Right hand draw X") emerges as the most possible to be the correct one with the passage of time. More extensive experimentation may lead to more robust findings concerning the similarity and relationship between classes.

\begin{table}
\caption{Confusion matrix of the proposed pipeline.}\label{tab1}
\begin{tabular}{cclcc|cccccccccccccccccccccccccccccccccccccccccccccccccc|ccc}
\multicolumn{58}{c}{\textbf{Truth}}\\
\hline
  &&  &&&&& a1 &&&&& a2 &&&&& a3 &&&&& a4 &&&&& a5 &&&&& a6 &&&&& a7 &&&&& a8 &&&&& a9 &&&&& a10 &&&&& Total \\ \hline
    \multirow{10}{*}{\rotatebox[origin=c]{90}{\textbf{Predicted}}}
&&a1 &&&&& \textbf{20} &&&&&	0  &&&&& 0  &&&&&	0  &&&&&  0 &&&&&	0  &&&&& 2  &&&&&	0  &&&&& 1  &&&&&	0   &&&&&	23 \\
&&a2 &&&&& 0	 &&&&& \textbf{10} &&&&& 1  &&&&& 0  &&&&&  0 &&&&& 0  &&&&& 0  &&&&& 0  &&&&& 0  &&&&&	0   &&&&&	11 \\
&&a3 &&&&& 0  &&&&&	12 &&&&& \textbf{26} &&&&&	0  &&&&&  0 &&&&&	0  &&&&& 0  &&&&&	1  &&&&& 0  &&&&&	0   &&&&&	39 \\
&&a4 &&&&& 0	 &&&&& 0  &&&&& 3  &&&&& \textbf{23} &&&&&  0 &&&&&	6  &&&&& 1  &&&&&	0  &&&&& 0  &&&&&	0   &&&&&	33 \\
&&a5 &&&&& 0	 &&&&& 0  &&&&& 0  &&&&&	5  &&&&& \textbf{30} &&&&& 0  &&&&& 0	 &&&&& 1  &&&&& 0	 &&&&& 0   &&&&&	36 \\
&&a6 &&&&& 4  &&&&&	0  &&&&& 0	 &&&&& 2  &&&&&  0 &&&&&	\textbf{24} &&&&& 2	 &&&&& 0  &&&&& 0  &&&&&	0   &&&&&	32 \\
&&a7 &&&&& 0	 &&&&& 0  &&&&& 0	 &&&&& 0  &&&&&  0 &&&&&  0 &&&&& \textbf{14} &&&&& 1  &&&&& 0  &&&&&	0   &&&&&	15 \\
&&a8 &&&&& 0	 &&&&& 7  &&&&& 0	 &&&&& 0  &&&&&  0 &&&&&  0 &&&&& 6	 &&&&& \textbf{27} &&&&& 0  &&&&& 0   &&&&&	40 \\
&&a9 &&&&& 6	 &&&&& 1  &&&&& 0	 &&&&& 0  &&&&&  0 &&&&&  0 &&&&& 5  &&&&& 0  &&&&& \textbf{29} &&&&& 0   &&&&&	41 \\
&&a10&&&&& 0	 &&&&& 0  &&&&& 0	 &&&&& 0  &&&&&  0 &&&&&  0 &&&&& 0	 &&&&& 0  &&&&& 0	 &&&&& \textbf{30}	&&&&&  30 \\ \hline
\end{tabular}
\end{table}

\section{Conclusion}
This paper presented a compact online action recognition methodology, based on a novel sequence encoding scheme and a lightweight convolutional neural network. It is suitable for deployment on hardware with limited capabilities, which can be found even in household robotic systems. Experimentation and cross validation showed this pipeline can be used with state-of-the-art human pose estimation techniques to form efficient end-to-end activity recognition techniques.
Discrepancies in the testing data pointed out that there may be room for improvement, as more novel convolutional neural network architectures emerge. Furthermore, there may be hints of overfitting, as stated in the previous section, which may need to be dealt with.

In future work, we will include attention models \cite{Rapantzikos2007}\cite{Wang_2017_CVPR} and temporal segmentation \cite{avr2000} in the CNN to further improve the performance of our approach. We will also focus on decrypting the inherent relationships and similarities between actions as encoded by $SRF$s, to evaluate the proposed scheme's performance as an action analysis and assessment tool.

%
%
%
\bibliographystyle{splncs04}
\bibliography{myreferences}
%




\end{document}